\documentclass[runningheads]{llncs}
\usepackage[]{eccv}

\usepackage{eccvabbrv}

\usepackage{graphicx}
\usepackage{booktabs}

\usepackage[accsupp]{axessibility}  

\usepackage[breaklinks,colorlinks,citecolor=eccvblue]{hyperref}

\usepackage{orcidlink}

\begin{document}

\title{\LARGE \bf {GraspXL: Generating Grasping Motions for Diverse Objects at Scale}}  

\author{Hui Zhang\inst{1} \and
Sammy Christen\inst{1} \and
Zicong Fan\inst{1, 2} \and
Otmar Hilliges\inst{1} \and
Jie Song\inst{1*}} 
\authorrunning{H. Zhang et al.}
\titlerunning{GraspXL}

\institute{ETH Z{\"u}rich, Switzerland \and
Max Planck Institute for Intelligent Systems, Germany}

\maketitle
\begin{figure}[t]
  \centering
  \includegraphics[width=1.0\linewidth]{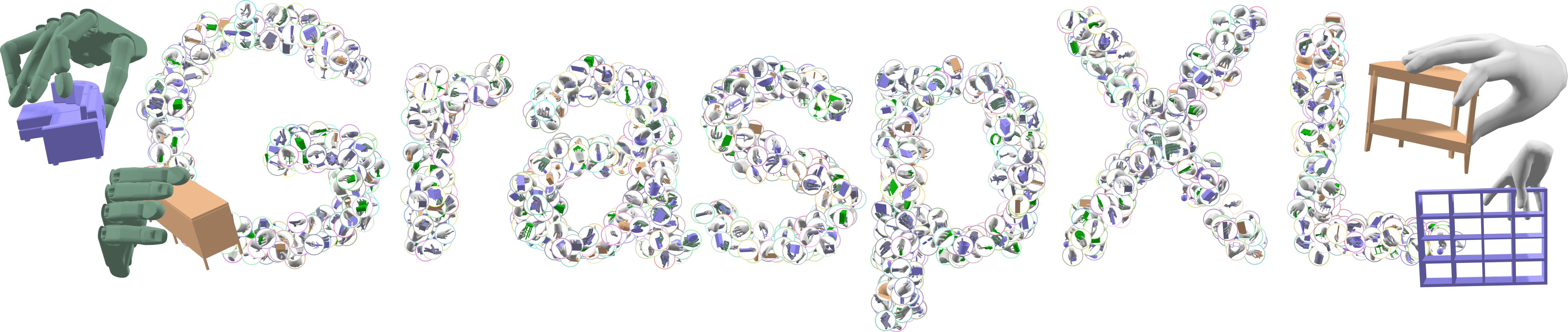}
  \caption{
  \textbf{Large-scale Grasping Synthesis}. 
    Our method, \method, can be used to generate large-scale grasps with robotic hands, and the MANO hand model. Here we show large-scale generated results, better viewed when zoomed in.
  }
\label{fig:tease_more}
\end{figure}
\begin{figure}[thpb]
  \centering
  \includegraphics[width=0.99\linewidth]{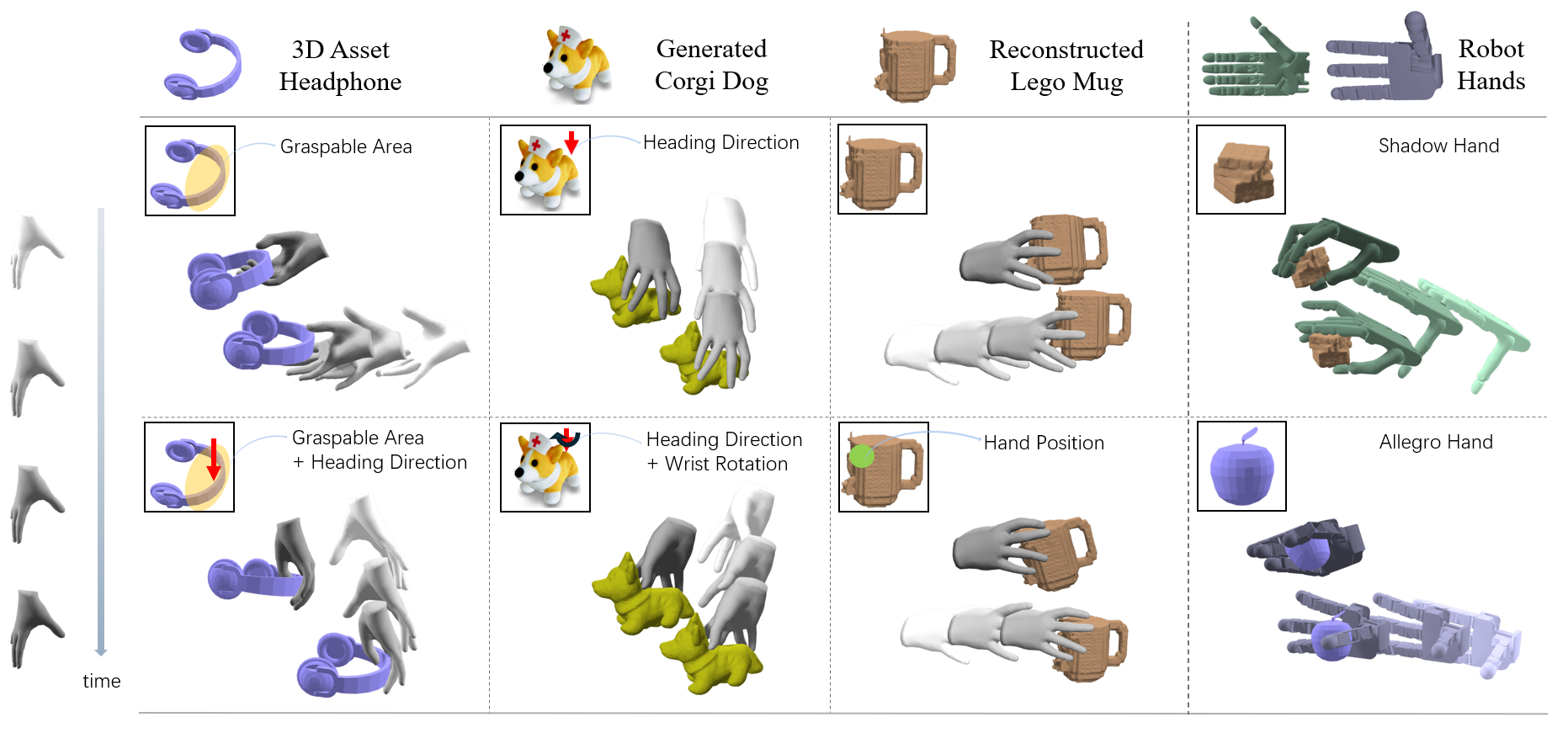}
  \caption{\textbf{\task}. Given a hand model and an object, our goal is to synthesize grasp motions that adhere to high-level objectives, which may consist of one or multiple objectives including graspable areas (indicated by the shadow), heading directions (indicated by the red arrow), wrist rotations (indicated by the black arrow), and positions of the hand (indicated by the green dot). For each sequence, the darker hand represents more recent in time.
  }
\label{fig:overview}
\end{figure}
\begin{abstract}
Human hands possess the dexterity to interact with diverse objects such as grasping specific parts of the objects and/or approaching them from desired directions. More importantly, humans can grasp objects of any shape without object-specific skills. 
Recent works synthesize grasping motions following single objectives such as a desired approach heading direction or a grasping area. Moreover, they usually rely on expensive \threeD hand-object data during training and inference, which limits their capability to synthesize grasping motions for unseen objects at scale. 
In this paper, we unify the generation of hand-object grasping motions across multiple motion objectives, diverse object shapes and dexterous hand morphologies in a policy learning framework \method. 
The objectives are composed of the graspable area, heading direction, wrist rotation, and hand position. 
Without requiring any \threeD hand-object interaction data, our policy trained with 58 objects can robustly synthesize diverse grasping motions for more than \textbf{500k} unseen objects with a success rate of 82.2\%. At the same time, the policy adheres to objectives, which  enables the generation of diverse grasps per object. Moreover, we show that our framework can be deployed to different dexterous hands and work with reconstructed or generated objects.
We quantitatively and qualitatively evaluate our method to show the efficacy of our approach. 
Our model, code, and the large-scale generated motions are available at \href{https://eth-ait.github.io/graspxl/}{https://eth-ait.github.io/graspxl/}.

\end{abstract}
\def\thefootnote{*}\footnotetext{Now at HKUST(GZ)\&HKUST}
\section{Introduction}
\label{introduction}

In our daily lives, we constantly engage with a wide variety of objects, from taking an apple out of a bowl and handling a knife by its grip to lifting a pair of headphones off the floor. This routine showcases the remarkable dexterity and adaptability of human hands, which effortlessly achieve complex tasks such as precisely grasping objects of different shapes in specific areas and from certain directions. Impressively, humans achieve this without needing object-specific pre-training, enabling us to manipulate items of any shape with ease. Therefore, the ability to generate versatile grasping motions that adhere to certain motion objectives -- like precise graspable areas and specific heading directions -- holds significant benefits for fields like animation and robotic grasping~\cite{ghosh2022imos, christen2023synh2r, agarwal2023dexterous}.

In this paper, we present \method, a policy learning framework capable of generating motions for a wide variety of objects, motion objectives, and hand morphologies, which is shown as \reffig{fig:overview}. 
Our method does not rely on any \threeD hand-object data to train but can robustly generalize to grasp a broad range of unseen objects. Consequently, our approach significantly scales hand-object motion generation, accommodating over \textbf{half a million} unseen objects, and we show some examples in \reffig{fig:tease_more}.
Existing methods for generating hand-object motions struggle with scalability to unseen objects due to their dependence on pose references \cite{zhang2024artigrasp, christen2022dgrasp}, their need for time-intensive optimization \cite{xu2023unidexgrasp, christen2023synh2r}, or their limitation to objects encountered during training \cite{mandikal2021dexvip}.
Furthermore, these existing approaches cannot, when applied off-the-shelf, generate interacting motions that fulfill multiple motion objectives.

Scaling objective-driven grasping motion generation to a wide variety of unseen objects poses several challenges. First, there is the \textbf{generalization ability}, where the learning framework should be general enough to handle different object shapes, dexterous hand models, and motion objectives. It is necessary to design a framework that avoids specific assumptions about the object shape and hand model. Second, the model must establish \textbf{stable grasping while adhering to multiple objectives}. Given the variety of object shapes and objectives, the model needs to learn stable grasping while satisfying multiple objectives. However, these high-level goals may negatively affect each other during training as the exploration of objectives can lead the model into local optima. In such cases, the objectives are followed initially, but no stable grasp is reached due to the object movement caused by contact, making the learning difficult and increasing the requirements for control precision.

We formulate \method in the reinforcement learning paradigm and leverage physics simulation. 
To allow a policy to react to varying object shapes, we capture the general shape features of diverse objects with the vectors from each finger joint to the nearest point on the object surface. To handle multiple objectives, we introduce a control scheme, dubbed \emph{objective-driven guidance}, that guides the hand towards the desired objective(s). To achieve generalization across diverse hands, we propose a general reward function composed of a grasping reward term and an objective reward term that is agnostic to the hand morphology.
Finally, to tackle the difficulty of learning stable grasping while satisfying the target objectives,
we propose a learning curriculum to decompose the learning process to objective learning and grasp learning. 
Specifically, we first train the policy on stationary objects with a larger objective reward to learn precise finger motions for the objectives. We then fine-tune the policy on non-stationary objects with a larger grasping reward to promote stable grasping. 

In our experiments, we evaluate methods for multi-objective grasping motion synthesis on PartNet~\cite{partnet2019} and ShapeNet~\cite{shapenet2015}. We enhance \synno~\cite{christen2023synh2r}, the only method offering controllability in hand heading direction, by incorporating more detailed motion objectives. Our approach, unlike \synno~\cite{christen2023synh2r}, achieves higher performance without time-intensive optimization for reference poses, yielding a 30\% increase in success rates and reducing objective errors by 30\%-50\%. Additionally, we demonstrate our method's broad applicability and generalization across a diverse range of conditions: it effectively handles over half a million objects from large-scale 3D datasets~\cite{objaverse}, adapts to objects from text-to-3D generation methods~\cite{poole2022dreamfusion}, applies to objects from 3D reconstruction techniques~\cite{fan2024hold}, and operates with various robotic hands, including Shadow~\cite{Shadow}, Allegro~\cite{Allegro}, and Faive~\cite{toshimitsu2023getting}. We validate our method's superiority over others through quantitative and qualitative measures and highlight its broad generalization capabilities.  Additionally, we dissect our framework's critical elements and investigate the impact of various objective combinations on performance.

In summary, our contributions are: 1) \method, a framework that synthesizes grasping motions on a large scale (500k+) of unseen objects, without relying on hand-object datasets during training. 2) A learning curriculum and objective-driven guidance to enable our method to achieve stable grasping while satisfying multiple objectives. 3) A dataset of diverse generated grasp motions for 500k+ objects with different hands. 4) We show that our method is general enough to be deployed on reconstructed or generated objects and different dexterous hands.
The code, models, and dataset are released on our project page.

\section{Related Work}
\label{related}

\begin{table}
\caption{
\textbf{Comparison with existing grasping motion synthesis methods.}
}
  \vspace{-4mm}
\label{tab:related_work}
\begin{center}
\resizebox{0.8\columnwidth}{!}{
\begin{tabular}{l|cccc}
   \toprule
\multirow{2}{*}{Method} & Multiple & Different & Data-agnostic & Number of Novel        \\
                        & Objectives & Hands     & Inference & Test Objects    \\
    \midrule
\dexvip          & \redcross & \redcross   & \redcross  & 0   \\
\dgrasp          & \redcross   & \redcross & \redcross   & 3    \\
\unidexgrasp     & \redcross   & \redcross & \greencheck & 100   \\
\unidexgraspplus & \redcross   & \redcross & \greencheck & 100    \\
\syn             & \redcross   & \redcross & \greencheck & 1,174   \\
\textbf{\methodname~(Ours)} & \greencheck & \greencheck & \greencheck & \textbf{503,409}   \\
   \bottomrule
\end{tabular}
}

\end{center}
\end{table}
We categorize related works into hand-object interaction synthesis and dexterous robot hand manipulation. \reftab{tab:related_work} compares different grasping motion synthesis methods with ours regarding the use of objectives, different hand morphologies, whether they require datapoints at inference time, and the number of reported results on unseen objects.

\subsection{Hand-object Interaction Synthesis}
In the literature, hand interaction primarily focuses on  hand(-object) reconstruction~\cite{tekin2019ho,fan2023arctic,liu2021semi,cao2021handobject,yang2021cpf,fan2021digit,duran2024hmp,ziani2022tempclr,fan2024benchmarks}, static grasp synthesis \cite{corona2020ganhand, turpin2023fastgraspd, jiang2021graspTTA, ye2023affordance} and temporal hand-object motion synthesis~\cite{zheng2023cams, christen2022dgrasp, braun2023physically, taheri2021goal, ghosh2022imos, zhang2021manipnet}. This work concentrates on the latter. In synthesis, some previous methods use pure data-driven approaches to synthesize human hand manipulation sequences and rely on post-processing to enhance physical plausibility \cite{zheng2023cams, zhang2021manipnet}. Data-driven methods are supervised by  3D hand-object annotated sequences during training~\cite{zheng2023cams, ghosh2022imos, taheri2021goal}, and some methods rely on references like wrist trajectories~\cite{zhang2021manipnet} during inference. However, acquiring accurate 3D hand-object data is infeasible to scale because of expensive capture setups~\cite{fan2023arctic,grab}. Due to limited training data, data-driven methods are constrained by their training distribution, making generalization challenging, especially for conditional generation tasks with motion objectives. Moreover, their dependence on data restricts their applicability across different dexterous hand platforms such as Shadow Hand \cite{Shadow} and Allegro Hand \cite{Allegro}. 

Some works leverage physics simulation within reinforcement-learning frameworks to alleviate the data requirement and ensure physical feasibility. Christen \etal~\cite{christen2022dgrasp} generate natural grasp sequences from captured or reconstructed static grasp references. Zhang \etal~\cite{zhang2024artigrasp} achieve two-hand grasp and articulation with a single frame reference grasp. However, both methods lack extensive evaluation to demonstrate their generalization ability, and their reliance on references restricts them from scaling to more objects. Some methods~\cite{christen2023synh2r, xu2023unidexgrasp} first generate grasp reference poses and then synthesize grasping motions accordingly for thousands of objects. However, they either rely on a time-consuming optimization process~\cite{christen2023synh2r, xu2023unidexgrasp} or are limited by the diversity in the pre-collected dataset~\cite{xu2023unidexgrasp}. Furthermore, the pre-defined references may not be physically feasible, which introduces extra disturbances to the generated motions as the hands should adhere strictly to the imperfect references. In contrast to existing approaches, our method does not require grasping references and offers real-time inference capabilities for grasping a wide range of objects, while at the same time enabling the control over multiple motion objectives such as graspable areas, heading directions, wrist rotations, and hand positions. 

\subsection{Dexterous Robot Hand Manipulation}
Dexterous manipulation plays a crucial role in enhancing robot capabilities \cite{Ze2023HInDex, ye2023learning, li2020mobile}. With the motivation to learn from humans, some approaches use imitation learning. However, this requires full human demonstrations for both training and inference \cite{qin2022dexmv, chen2022dextransfer, liu2023dexrepnet}, which are usually expensive to collect. Some other methods utilize retargeted human demonstrations during training \cite{ye2023learning}, use teleoperated sequences as training data \cite{rajeswaran2018learning}, or learn a parameterized reward function based on demonstrations~\cite{christen2019hri}. However, the reliance on expensive full grasping sequences during training still limits their generalization ability for out-of-distribution settings. Instead of relying on full trajectories, Xu \etal~\cite{xu2023unidexgrasp} use static pose references to guide the motion by first predicting the contact map and then optimizing the poses accordingly, with physics-based heuristics used to filter out invalid poses. 
This process is expensive and makes real-time inference infeasible. In contrast, our method generates motions in real-time while adhering to fine-grained motion objectives. Furthermore, without reliance on hand-object interaction data, our method can easily scale to 500k objects.

Another line of research learns grasping with RL without conditioning on reference grasps~\cite{chen2022humanlevel, ding2023learning, qin2023dexpoint}. Wan \etal~\cite{Wan_2023_ICCV} propose a curriculum learning framework to distill a state-based policy into a vision-based policy. They generalize to many objects and achieve impressive results on vision-based grasping, whereas we focus on generalizable grasping with different objectives in the state-based setting. Similar to ours, some recent methods can achieve affordance-aware grasps. Mandikal \etal~\cite{mandikal2020graff} train an affordance-aware grasp policy with RL and evaluated with only 24 novel objects, and further introduce a hand pose prior from YouTube videos for natural hand configuration~\cite{mandikal2021dexvip} which is only evaluated with objects used for training. Agarwal \etal~\cite{agarwal2023dexterous} learn a category-level policy that grasps the objects by affordance areas, but they are limited to grasp their training categories and can generalize to only 1 unseen category. Moreover, all of these methods only generate a single pose per object. In contrast, our method can generate diverse affordance-aware grasping poses that can be controlled via motion objectives, and can be deployed for over 500k unseen objects.
\begin{figure*}[thpb]
  \centering
  \includegraphics[width=0.9\textwidth]{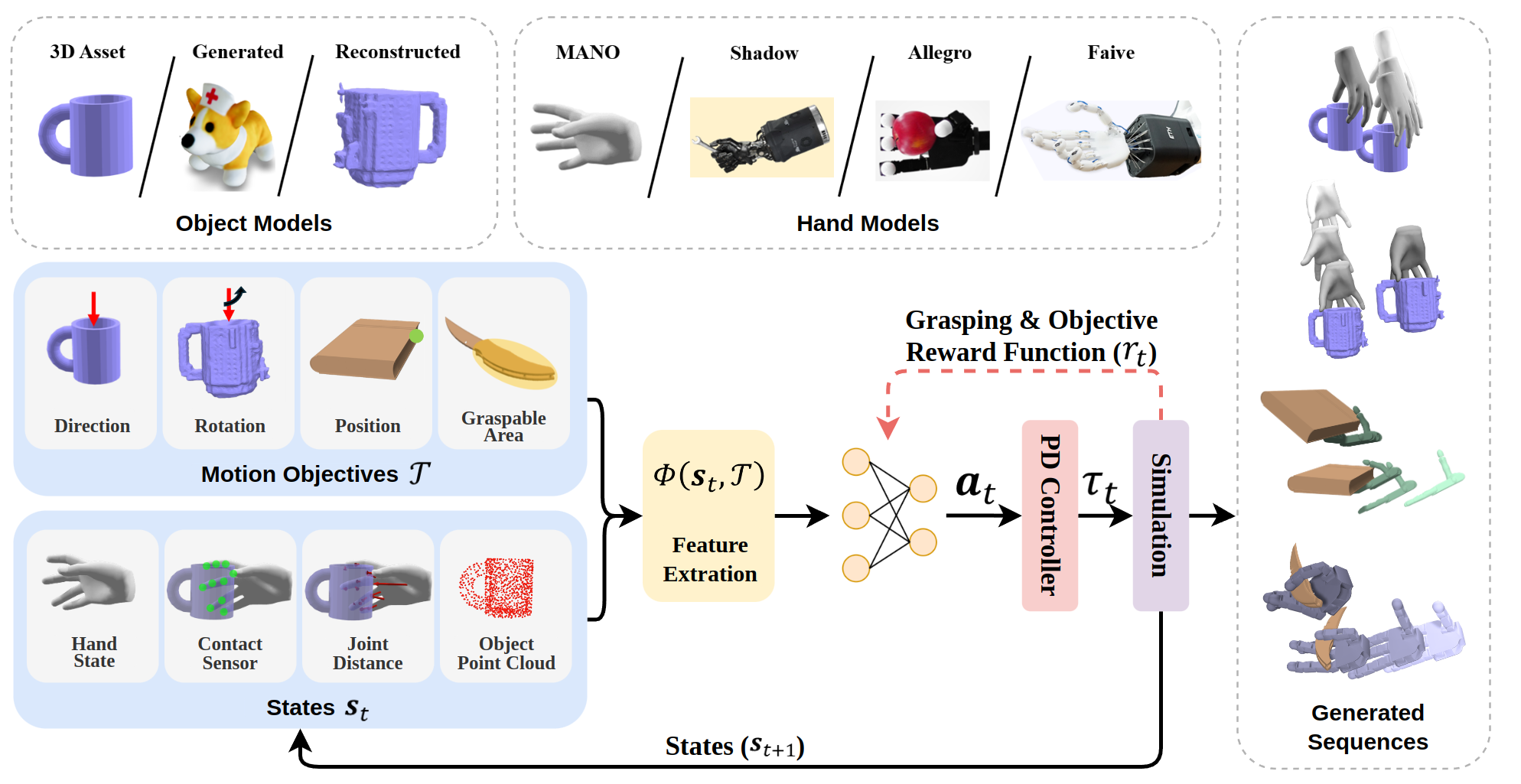}
  \caption{\textbf{Overview of \method}. As shown in the top row, our method can utilize captured, generated, or reconstructed objects, and different dexterous hand platforms such as MANO, Shadow, Allegro or Faive. With given object and hand model, the policy takes different objectives and states as inputs (on the left), and outputs dynamic grasp motions according to the specific objectives (on the right, accordingly, where darker hand represents more recent in time). The objectives can be the heading direction, wrist rotation, hand position or graspable area, and the states contain the hand state, contact, force and distance of each link with the object, and the object point cloud.}
    \vspace{-4mm}
\label{fig:pipeline}
\end{figure*}

\section{GraspXL}
\begin{figure}[thpb]
  \centering
  \includegraphics[width=1.0\linewidth]{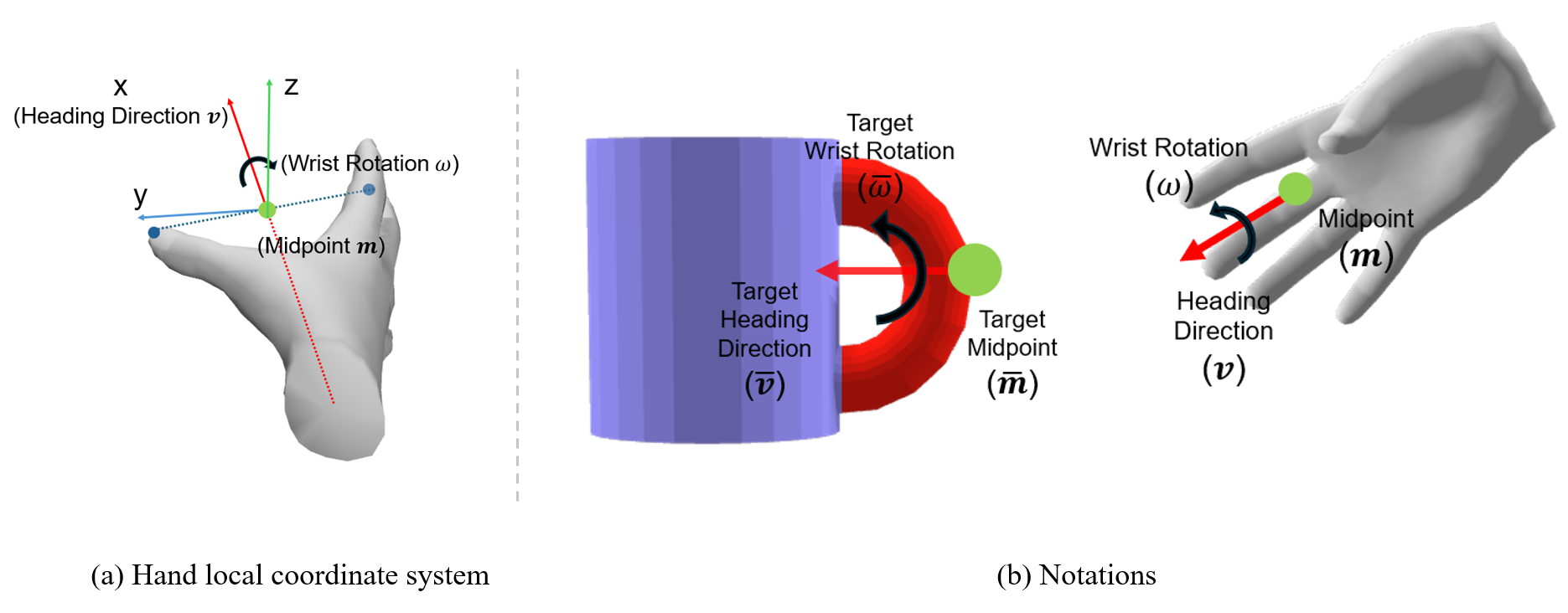}
  \caption{\textbf{Task definition.} 
  (a) The local coordinate system of the hand where the x-axis is the heading direction $\V{v}$, the origin is the midpoint position $\V{m}$ (see text for definition), the rotation about $\V{v}$ is $\omega$. 
  (b) 
  Given an object with user-specified graspable $\{\V{o}^+_j\}$ and non-graspable vertices $\{\V{o}^-_j\}$ (labelled in red and blue),
  the goal of the agent is to approach and grasp the object while satisfying motion objectives $\bar{\V{v}}$, $\bar{\V{m}}$, $\bar{\omega}$, and contact with the graspable area $\{\V{o}^+_j\}$.
  }
\label{fig:frame}
\end{figure}

\myparagraph{Task Definition}
\label{sec:def}
As illustrated in \reffig{fig:frame}, we assume a hand model $h$ with $L$ links where $\V{h}_i\in \R^3$ is the position of the i-th link. The hand pose consists of joint angles $\V{q}\in \R^{L\times 3}$ and the global orientation represented by the heading direction $\V{v}\in \R^3$ and wrist rotation $\omega \in \R$ about the directional vector $\V{v}$. As the hand moves, it has linear and angular velocities $\V{u}_h\in \R^6$. 
We define  $\V{m\in \R^3}$ as the midpoint between the thumb tip and the third joint of the middle finger.
We also assume a rigid object $o$ which has a point cloud of \threeD vertices $\{\V{o}_j\}$. A user may partition the point cloud into graspable/non-graspable points, specifically $\{\V{o}_j\} = \{\V{o}^+_j\} \cup \{\V{o}^-_j\}$, where we split the points into two disjoint sets to specify whether the point should be encouraged to be in contact with the hand or not when motion is generated.
As the object moves, it also has linear and angular velocities $\V{u}_o\in \R^6$. 
The hand links may be in contact $\V{c}\in \{0, 1\}^{L}$ with the object using forces in magnitude $\V{f} \in \R^{L}$. 
In particular, the links may contact with the graspable/non-graspable area, which are denoted as $\V{c}^{+}\in \{0, 1\}^{L}$  an $\V{c}^{-}\in \{0, 1\}^{L}$. Similarly, the force magnitude vector $\V{f}$ can be decomposed into $\V{f}^+ \in \R^{L}$ and $\V{f}^- \in \R^{L}$.

A user may specify motion objectives $\mathcal{T}$ to define the target heading direction $\bar{\V{v}}$, wrist rotation $\bar{\omega}$, midpoint position $\bar{\V{m}}$, the partition of graspable/non-graspable object point cloud $\{\V{o}^+_j\} \cup \{\V{o}^-_j\}$. 
Given this specification, our goal is to generate a motion sequence that approaches and grasps an object without dropping whilst adhering to the objectives.
Note that only the target heading direction $\bar{\V{v}}$ is mandatory to specify in $\mathcal{T}$. 
If not specified, $\{\V{o}_j\}$ will equal to $\{\V{o}^+_j\}$, $\bar{\V{m}}$ will be the mean of $\{\V{o}_j\}$, $\bar{\omega}$ will be zero. See \suppl for more details.

\myparagraph{Overview}
\reffig{fig:pipeline} outlines our RL-based method, \method. 
Given the objectives $\mathcal{T}$ and the state $\textbf{s}$ obtained from the physics simulation, we utilize a feature extraction layer $\Phi$ to derive the features. Subsequently, the policy $\policyvec$ takes features as input and generates the actions $\textbf{a}$, representing the PD-control targets used to compute the torques $\torques$. These torques are then applied to the joints of the hand model in the physics simulation to update the state $\textbf{s}$, which is subsequently fed back into our feature extraction layer for the next iteration.\\

\myparagraph{Reinforcement Learning Background}
The task is formulated as a standard Markov Decision Process (MDP). The goal is to determine the policy $\policyvec$ that maximizes the expected reward $\mathbb{E}_{\xi \sim \policyvec} \left[ \sum_{t=0}^T\mathcal{\gamma}^t r_t \right]$, where $\mathcal{\gamma} \in [0, 1]$ denotes the discount factor, $r_t$ represents the reward at time step $t$, and $\mathcal{\xi}=[(\states_0, \actions_0), \cdots, (\states_T, \actions_T)]$ denotes a state-action pair trajectory generated by the policy $\policyvec$ interacting with the physics simulation. 
The trajectory $\mathcal{\xi}$ is determined by the transition function $p(\states_{t+1}|\states_t, \actions_t)$, which is governed by the physics simulation along with an initial state distribution $p(s_0)$. The distribution of a trajectory $\mathcal{\xi}$ is defined as $ p_{\theta}(\mathbf{\xi})=p(\states_0)\prod_{t=0}^T p(\states_{t+1}|\states_t, \actions_t) \policyvec(\actions_t| \Phi(\textbf{s}_t,\mathcal{T}))$. 
Here, $\policyvec$ is represented by a neural network. \\

\myparagraph{Feature Extraction}
Given a hand-object state $\textbf{s}$ and the motion objectives $\mathcal{T}$, 
we extract the following features with a conversion function $\Phi(\textbf{s},\mathcal{T})$:
\begin{equation}
    \Phi(\textbf{s},\mathcal{T}) = (\textbf{q}, \textbf{d}, \V{u}_h, \V{u}_o, \textbf{c}, \textbf{f}, \Tilde{\textbf{v}}, \Tilde{\textbf{m}}, \Tilde{\omega}, \textbf{l}),
\end{equation}
\noindent which includes finger joint angles $\textbf{q}$, the finger joint tracking error (compared with the target angles for PD controller) $\textbf{d}$,  the hand and object velocities $\V{u}_h$ and $\V{u}_o$, the contact vector $\textbf{c}$, the force magnitude vector $\textbf{f}$. 
Finally, $\Tilde{\textbf{v}}, \Tilde{\textbf{m}}, \Tilde{\omega}$ represent the differences between current and target heading directions, midpoint positions, and wrist rotation angles.
To represent shape features of an object and help the hand get aware of the graspable/non-graspable areas, we construct distance features $\textbf{l}^+\in \R^{L\times 3}$ where each row is the difference vector between a link position $\V{h}_i$ to the closet object vertex in the graspable part $\{\V{o}^+_j\}$. 
Similarly, we construct $\textbf{l}^{-}\in \R^{L\times 3}$ for the non-graspable part $\{\V{o}^-_j\}$.
We ablate this component in~\refsec{ablation}. \\

\myparagraph{Reward Function}
The reward function should guide our policy to learn a solution that can grasp the objects with the desired objectives while at the same time achieving successful grasps. Furthermore, it should be formulated without specific assumptions about the hand morphology so as to be applied on different hand models. As a result, we define our reward function as follows:
\begin{equation}
    r = r_{\text{goal}} + r_{\text{grasp}},
\end{equation}
where $r_{\text{goal}}$ is for motion objectives and $r_{\text{grasp}}$ is for successful grasping.

In particular, the motion objective reward $r_{\text{goal}}$ is formulated as:
\begin{equation}
r_{\text{goal}} = r_{\text{dis}} + r_{\V{v}} + r_{\omega} + r_{\V{m}}
\end{equation}
where $r_{\text{dis}}$ promotes approaching the target while avoiding non-graspable areas, and $r_{\V{v}}$, $r_{\omega}$, and $r_{\V{m}}$ reward aligning heading direction, wrist rotation, and midpoint, respectively.
Concretely, the term $r_{\text{dis}}$ is defined as
\begin{equation}
\scalebox{0.9}{
   $r_{\text{dis}} = -\sum_{i=1}^L \left[w_{d}^{+}(i)||\textbf{h}_i-\textbf{o}^{+}_i||^2 - w_{d}^{-}(i)||\textbf{h}_i-\textbf{o}^{-}_i||^2\right]$,
}
\end{equation}
with the weights $w_{d}^{+}(i)\in\R$ and $w_{d}^{-}(i)\in\R$, the i-th link position ($\textbf{h}_i$), the closest graspable/non-graspable object points to the i-th link ($\textbf{o}^{+}_i$ and $\textbf{o}^{-}_i$). 
The rewards $r_{\V{v}}$, $r_{\omega}$, and $r_{\V{m}}$ penalize discrepancies in the heading direction $\textbf{v}$, the wrist rotation angle $\omega$, and midpoint position $\textbf{m}$ from the targets:
\begin{equation}
\scalebox{0.9}{
    $r_{\V{v}} = -w_{\V{v}}||\textbf{v}-\overline{\textbf{v}}||^2,\  
    r_{\omega} = -w_{\omega}||\omega-\overline{\omega}||^2,\ 
    r_{\V{m}} = -w_{\V{m}}||\textbf{m}-\overline{\textbf{m}}||^2$.
}
\end{equation}

The grasp reward $r_{\text{grasp}}$ promotes proper contact and natural poses by combining multiple factors:
\begin{equation}
r_{\text{grasp}} = r_{\V{c}} + r_{\V{f}} + r_{\text{anatomy}} + r_{\text{reg}}
\end{equation}
where $r_{\V{c}}$ assesses the contact between fingers and the object, considering both the target and non-target areas ($r_{\V{c}} = w_{c}^+\norm{\textbf{c}^{+}}^2 - w_{c}^{-}\norm{\textbf{c}^{-}}^2$). The force term, $r_{\V{f}}$, encourages the contact forces with the graspable area and punishes the contact forces with the non-graspable area, capped by a factor proportional to the object’s weight $w_o$: ($r_{\V{f}} = w_{f}^+\textbf{c}^{+}\cdot min(\textbf{f}^{+}, \lambda w_o) - w_{f}^-\textbf{c}^{-}\cdot min(\textbf{f}^{-}, \lambda w_o)$). $r_{\text{reg}}$ penalizes excessive velocities to ensure stability ($r_{\text{reg}} = -w_{h}||\V{u}_h||^2 - w_{o}||\V{u}_o||^2$). 
If the hand model is \mano~\cite{MANO:SIGGRAPHASIA:2017}, we additionally apply the anatomy reward $r_{\text{anatomy}}$~\cite{yang2021cpf} on joint angles $\V{q}$ for natural poses. \\

\myparagraph{Curriculum}
Generating grasping motion with objectives requires a policy that establishes stable grasps while achieving specific goals. This is complicated because of the potential for adverse outcomes like object flipping caused by contact when trying to accomplish the objectives. To mitigate this, we introduce a learning curriculum: we start by training the policy on static objects with increased $r_{\text{goal}}$ to hone precise finger movements for objectives. Training progresses to moving objects with a higher $r_{\text{grasp}}$, enhancing wrist movements for secure, non-slip grasps. The effectiveness of this approach is discussed in~\refsec{ablation}. \\

\myparagraph{Objective-driven hand guidance}
Diverse objectives complicate policy exploration due to the need for varied wrist movements. To address this, we present a simple-yet-effective method to guide the hand during training and inference, improving exploration and control precision.
Essentially, we compute the differences between the target and the current values for the heading direction $\V{v}$, the wrist rotation angle $\omega$, and the midpoint position $\V{m}$. These differences are then applied directly as bias terms for the wrist's 6 degrees of freedom (DoF) PD-controller to guide the wrist toward motion objectives. 
This simple trick promotes quicker convergence and boosts performance, as detailed in \refsec{ablation}.

\section{Experiments}
\label{sec:experiments}

This section involves several experiments to evaluate our method's effectiveness and generalization capabilities. We detail the experimental setups in \refsec{experiment}, compare our method's performance against others in \refsec{evaluation}, and examine its generalization across various unseen objects and hand models in \refsec{generalization}. Lastly, we ablate our method's components and analyze the influence of different objective combinations in \refsec{ablation}.

\subsection{Experimental Setup}
\label{experiment}

\myparagraph{Datasets}
To construct the training set, we randomly select 26 objects and 32 objects from ShapeNet~\cite{shapenet2015} and PartNet~\cite{partnet2019} respectively. 
To demonstrate our method's generalization to large-scale object datasets, we use Objaverse~\cite{objaverse}.
Since not all objects in the three datasets are suitable for rigid-body grasping (\eg a piece of paper), we filter out objects with such shapes. 
After filtering and preprocessing, we have 48, 3993 and 503k objects in PartNet, ShapeNet, and Objaverse for testing, respectively. The PartNet test set contains unseen objects from seen categories, while the objects of the other two test sets are completely novel.
Please see \suppl for details on dataset filtering and preprocessing.

\myparagraph{Training Details}
\label{detail}
We use PPO~\cite{schulman2017proximal} for RL training with the RaiSim~\cite{Hwangbo2018Raisim} physics simulator. Experiments are conducted using a single Nvidia RTX 6000 GPU with 128 CPU cores. To improve data diversity for better generalization, 
we sample objectives during training with the following heuristics: we randomly sample the target heading direction $\bar{\V{v}}\in \R^3$ and wrist rotation angle $\bar{\omega}\in [0, 2\pi)$ while ensuring that the graspable part $\{\V{o}^+_j\}$ of the object is narrower than 12 cm between the thumb and the other fingers (along the y axis of the hand local coordinate in \reffig{fig:frame}a).
As a result, the object will not be too large to be grasped. To find a midpoint, we randomly sample a point on the graspable surface $\{\V{o}^+_j\}$.

\myparagraph{Evaluation Protocol} 
We measure the stability of the grasps with a success metric metric. To assess the adherence to the objectives, we compute several metrics and utilize the ShapeNet and PartNet (which contains part-based objects) datasets. \textbf{Midpoint Error (Mid. Error):} The mean Euclidean distance between the final midpoint position $\V{m}$ and the target $\bar{\V{m}}$ measured in centimeters. \textbf{Heading Error (Head. Error):} The mean geodesic distance between the final heading direction $\V{v}$ and the target $\bar{\V{v}}$ measured in radian. \textbf{Wrist Rotation Error (Rot. Error):} The mean absolute error between the final wrist rotation $\omega$ and the target $\bar{\omega}$ measured in radian. textbf{Contact Ratio:} The ratio between the number of links contacted with the graspable part $\{\V{o}^+_j\}$ and the entire object $\{\V{o}_j\}$. \textbf{Grasping Success Rate (Suc. Rate):} A grasp is determined as a success if the object is lifted higher than 10cm and remains stable without falling until the sequence terminates.

\myparagraph{Baselines} 
We choose \synno as the main baseline, because it offers some degree of controllability in the motion generation, (in terms of heading direction $\V{v}$). We adapt it with more fine-grained motion objectives. In particular, we have the following baselines. \textbf{\synno}: The original method generates grasping motion by first generating static grasping reference poses with an optimization procedure. Then it uses an RL-based policy to approach and follow the grasping reference pose. To adapt this method for objective-driven synthesis beyond heading direction $\V{v}$, we add additional optimization terms to include the wrist rotation error $\omega$ and the midpoint error $\V{m}$. We also encourage the contacts with the graspable part while punishing the contacts with the non-graspable part during the optimization. The RL-based policy stays the same as the original one in \synno, which is trained with the generated grasping references. \textbf{\synnopd}: The grasping references are generated the same way as in \synno above. We then set the reference poses as targets for the PD controller following \cite{christen2022dgrasp}.

\subsection{Method Comparison}
\label{evaluation}
Following our evaluation protocol in \refsec{sec:experiments},  we compare with \synno and \synno-PD for objective-driven motion synthesis. All methods use the \mano hand model and the same pre-sampled objectives.
We initialize the hand state in the same way for all methods.
In particular, the starting hand pose is set to the mean pose of \mano with an open thumb (see \reffig{fig:frame}a), and the wrist is positioned 30$cm$ away from the target midpoint position $\bar{\V{m}}$ along the target heading direction $\bar{\V{v}}$. For each sequence, we control the hand to grasp the object, and then apply torques to the wrist to lift the object. 
It's noteworthy that our method can directly infer in simulation whereas the baselines rely on pre-generated hand pose references through a time-consuming optimization procedure. 

\begin{table}[t]
\centering
\caption{
\textbf{Method Comparison on PartNet and ShapeNet.}
}
\label{tab:main}
\resizebox{\columnwidth}{!}{
\begin{tabular}{l|ccccc|cccc}
   \toprule
   & \multicolumn{5}{c|}{PartNet Test Set} & \multicolumn{4}{c}{ShapeNet Test Set} \\
\midrule
\multirow{2}{*}{Method} & Suc. Rate  & Mid. Error   & Head. Error    & Rot. Error   & Contact Ratio  & Suc. Rate  & Mid. Error   & Head. Error    & Rot. Error  \\
         & [\%] $\uparrow$ & [cm] $\downarrow$ & [rad] $\downarrow$ & [rad] $\downarrow$ & [\%] $\uparrow$ & [\%] $\uparrow$ & [cm] $\downarrow$ & [rad] $\downarrow$ & [rad] $\downarrow$ \\
\midrule
PD & 26.5 & 4.30 & 0.767 & 0.857 & 13.0 & 21.9 & 4.60 & 0.850 & 0.964 \\
SynH2R & 82.3 & 4.06 & 0.522 & 0.568 & 53.4 & 65.8 & 4.49 & 0.642 & 0.688 \\
Ours & \textbf{95.0} & \textbf{2.85} & \textbf{0.270} & \textbf{0.306} & \textbf{86.7} & \textbf{81.0} & \textbf{3.22} & \textbf{0.292} & \textbf{0.338} \\
   \bottomrule
\end{tabular}
}
\end{table}

\myparagraph{PartNet Evaluation}
\label{affordance}
We use the PartNet test set to evaluate all methods with the given objectives. For each object, we calculate the average performance among 25 randomly sampled heading directions $\bar{\V{v}}$, wrist rotations $\bar{\omega}$ and midpoints $\bar{\V{m}}$.
\reftab{tab:main} shows that our method significantly outperforms the baselines across all metrics. 
In particular, our method more closely follows the objectives in terms of hand grasping position (Midpoint Error), heading direction (Heading Error), wrist rotation (Wrist Rotation Error), and contact points with the graspable/non-graspable parts (Contact Ratio). At the same time, we achieve the most stable grasps (Grasping Success Rate). 

\myparagraph{ShapeNet Evaluation}
\label{large}
While PartNet specializes in part-based objects at a smaller scale, we extend our evaluation to ShapeNet to test on more diverse and unseen objects. We measure average performance across 5 randomly sampled heading directions $\bar{\V{v}}$, wrist rotations $\bar{\omega}$ and midpoints $\bar{\V{m}}$ per object. As shown in \reftab{tab:main}, our method outperforms all baselines, demonstrating less decrease in performance when comparing PartNet and ShapeNet, thereby proving superior generalization. 
The Grasping Success Rate has a drop for all the methods (even for the non-learning-based \synnopd baseline), which we claim is caused by some objects that are too large or too heavy to be grasped. It is worth noting that generating grasp pose references for the ShapeNet test set using the baselines requires approximately a week, whereas our method can directly perform real-time inference in simulation, showcasing its efficiency.
Please refer to \suppl for further qualitative comparisons.

\subsection{Generalization}
\label{generalization}
\myparagraph{Generalization to Large-scale Object Dataset}
Our method's scalability was tested using the large Objaverse~\cite{objaverse} dataset, resulting in a test set of over half a million objects of three different sizes: small, medium and large (refer to \suppl for preprocessing details). Performance on this set, as detailed in \reftab{tab:universal}, is comparable to our ShapeNet results, underscoring our method's ability to scale. Specifically, medium-sized objects offer optimal grasping success. Smaller objects enable higher success rates and more accurate grasp positioning but suffer from larger heading direction and wrist rotation errors, as smaller objects are easier to grasp but also easier to rotate in hand.
Overall, our method consistently shows excellent performance across different object sizes, confirming its effectiveness for varying scales. The generated grasping motions of different hands are released for further research.

\begin{table}[t]
\centering
\caption{
\textbf{Generalization to the Large-scale Objaverse Dataset.}
}
\vspace{-1mm}
\label{tab:universal}
\resizebox{0.8\columnwidth}{!}{
\begin{tabular}{l|cccc}
   \toprule
Objects & Suc. Rate [\%] $\uparrow$  & Mid. Error [cm] $\downarrow$  & Head. Error [rad] $\downarrow$   & Rot. Error [rad] $\downarrow$ \\
\midrule
Small  & \textbf{85.9} & 3.20 & 0.311 & 0.362  \\
Medium & 84.5 & \textbf{3.16} & 0.274 & 0.315 \\
Large  & 79.0 & 3.50 & \textbf{0.271} & \textbf{0.306} \\
\midrule
Average & 82.2 & 3.32 & 0.279 & 0.319 \\
   \bottomrule
\end{tabular}
}
\end{table}

\begin{table}[t]
\centering
\caption{
\textbf{Generalization to Generated and Reconstructed Objects.}
}
\vspace{-1mm}
\label{tab:recon}
\resizebox{\columnwidth}{!}{
\begin{tabular}{l|cccc}
   \toprule
Objects & Suc. Rate [\%] $\uparrow$  & Mid. Error [cm] $\downarrow$  & Head. Error [rad] $\downarrow$   & Rot. Error [rad] $\downarrow$ \\
\midrule
Generated Obj.     & 88.4 & 2.85 & 0.310 & 0.361  \\
Reconstructed Obj. (in the wild) & 77.5 & 3.68 & 0.222 &  0.281 \\
Reconstructed Obj. (YCB)  & 74.0 & 3.84 & 0.207 & 0.247 \\
Ground-truth Obj. (YCB)  & 73.7 & 3.63 & 0.259 & 0.301  \\
   \bottomrule
\end{tabular}
% \vspace{-7mm}
}
\vspace{-3mm}
\end{table}

\myparagraph{Generalization to Reconstructed and Generated Objects}
Our method not only synthesizes grasping motions for standard 3D assets \cite{objaverse,partnet2019,shapenet2015} but also effectively handles reconstructed and generated objects, expanding its usability. We evaluated its performance on objects reconstructed via HOLD~\cite{fan2024hold} and objects generated by DreamFusion~\cite{poole2022dreamfusion}. We analyzed the average performance for each object across 25 random sets of motion objectives, with findings detailed in \reftab{tab:recon}. Despite the objects being novel and containing severe artifacts (see the Rubik's Cube in \reffig{fig:overview}), our method maintains similar performance across all metrics compared to other experiments. Notably, the performance of objects reconstructed from HO3D~\cite{hampali2020ho3d} is similar with the performance of ground-truth objects. 
This shows our method's robustness towards reconstruction noise in object meshes. Please refer to the \suppl for a more detailed setting explanation.

\begin{table}[t]
\centering
\caption{
\textbf{Generalization to Different Hand Models.}
}
\vspace{-1mm}
\label{tab:different}
\resizebox{\columnwidth}{!}{
\begin{tabular}{l|ccccc|cccc}
   \toprule
   & \multicolumn{5}{c|}{PartNet Test Set} & \multicolumn{4}{c}{ShapeNet Test Set} \\
\midrule
Hand & Suc. Rate  & Mid. Error   & Head. Error    & Rot. Error   & Contact Ratio  & Suc. Rate  & Mid. Error   & Head. Error    & Rot. Error  \\
Model & [\%] $\uparrow$ & [cm] $\downarrow$ & [rad] $\downarrow$ & [rad] $\downarrow$ & [\%] $\uparrow$ & [\%] $\uparrow$ & [cm] $\downarrow$ & [rad] $\downarrow$ & [rad] $\downarrow$ \\
\midrule
Allegro & 95.3 & 4.38 & 0.291 & 0.300 & 81.1 & 83.4 & 4.45 & 0.271 & 0.292 \\
Shadow & 94.0 & 3.57 & 0.317 & 0.320 & 83.4 & 83.2 & 3.67 & 0.363 & 0.381 \\
Faive & 95.8 & 2.85 & 0.228 & 0.243 & 88.7 & 82.4 & 3.44 & 0.250 & 0.262 \\
\midrule
MANO & 95.0 & 2.85 & 0.270 & 0.306 & 86.7 & 81.0 & 3.22 & 0.292 & 0.338 \\
   \bottomrule
\end{tabular}

}
\end{table}
\begin{figure}[t]
  \centering
  \includegraphics[width=0.99\linewidth]{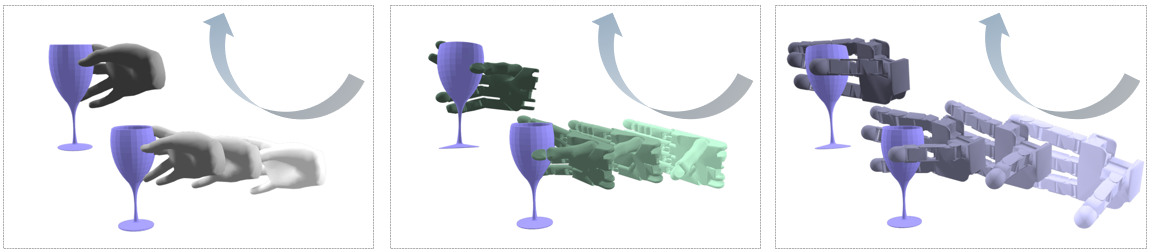}
  % \vspace{-1mm}
\caption{\textbf{Generated Motions of Different Hands with the Same Objectives.}
  We require the hands to approach from the right and grasp the upper part of the glass. 
  }
  \vspace{-3mm}
\label{fig:different}
\end{figure}

\myparagraph{Generalization on Different  Robotic Hands}
We assess our framework's generalization capabilities on various dexterous robotic hands, including Shadow Hand \cite{Shadow}, Allegro Hand \cite{Allegro}, and Faive Hand \cite{toshimitsu2023getting}, each differing in size and joint structure (Faive Hand has 30, Shadow Hand 22, Allegro Hand 16, and MANO 45 finger joints). Our method adapts seamlessly to different models by adjusting hand state and action space dimensions. According to \reftab{tab:different}, all tested hands showed similar success rates. However, the Allegro Hand has greater Mid. Errors and lower Contact Ratios due to its larger size, which makes it difficult to achieve precise position control. The Shadow Hand shows marginally higher objective errors due to its broad, flat palm which limits its dexterity. Despite these structural differences affecting performance metrics, all hands achieved commendable results, highlighting our method's ability to generalize. 
% \red{This suggests our framework's capacity for achieving human-like robotic grasping, whether following predefined or video-extracted human manipulation objectives.} 
This suggests our framework's capacity to deal with hand models with different morphologies.
\reffig{fig:different} illustrates the variety of grasping motions generated for different hands under the same motion objectives.

\subsection{Ablation and Analysis}
\myparagraph{Ablation}
\label{ablation}
We conduct an ablation study to evaluate the impact of different components in our method. Specifically, we assessed variations without the hand guidance technique (w/o Guidance), the joint distance features (w/o Distance), and the learning curriculum (w/o Curriculum). Results are detailed in \reftab{tab:ablation}.
The version without the curriculum slightly outperforms in contact ratio on the PartNet set and maintains similar success rates across tests but suffers significantly in objective-related metrics. We also notice the absence of the curriculum makes the results more sensitive to the exact reward coefficients, leading to more grasp failure with slightly higher objective rewards or poor objective precision with higher grasping rewards. These underscore the curriculum's role in decoupling the learning of stable grasping and objective fulfillment, which can help the policy avoid getting stuck in local optima caused by their influence on each other during training. 
The model without the joint distance features gets a slightly larger midpoint error and significantly underperforms in all other aspects, highlighting the distance features' value in understanding object shapes and adjusting to contact-induced movements.
Lastly, excluding hand guidance worsens all performance metrics, proving its efficacy in directing the hand to achieve the desired grasp. 

\begin{table}[t]
\centering
\caption{\textbf{Effects of Different Components in \method.}
}
\vspace{-1mm}
\label{tab:ablation}
\resizebox{\columnwidth}{!}{
\begin{tabular}{l|ccccc|cccc}
   \toprule
   & \multicolumn{5}{c|}{PartNet Test Set} & \multicolumn{4}{c}{ShapeNet Test Set} \\
\midrule
\multirow{2}{*}{Model} & Suc. Rate  & Mid. Error   & Head. Error    & Rot. Error   & Contact Ratio  & Suc. Rate  & Mid. Error   & Head. Error    & Rot. Error  \\
         & [\%] $\uparrow$ & [cm] $\downarrow$ & [rad] $\downarrow$ & [rad] $\downarrow$ & [\%] $\uparrow$ & [\%] $\uparrow$ & [cm] $\downarrow$ & [rad] $\downarrow$ & [rad] $\downarrow$ \\
\midrule
w/o Guidance & 90.0 & 3.22 & 0.394 & 0.425 & 82.2 & 68.5 & 3.74 & 0.455 & 0.528 \\
w/o Distance & 81.6 & 2.90 & 0.419 & 0.475 & 84.2 & 70.7 & 3.34 & 0.467 & 0.510 \\
w/o Curriculum & \textbf{96.2} & 4.12 & 0.381 & 0.462 & \textbf{88.8} & 79.6 & 4.60 & 0.396 & 0.461 \\
\midrule
Ours & 95.0 & \textbf{2.85} & \textbf{0.270} & \textbf{0.306} & 86.7 & \textbf{81.0} & \textbf{3.22} & \textbf{0.292} & \textbf{0.338} \\
   \bottomrule
\end{tabular}
}
\end{table}
\begin{table}[t]
\centering
\caption{\textbf{Generation Performance with Different Objective Combinations.}
}
\vspace{-1mm}
\label{tab:individual}
\resizebox{\columnwidth}{!}{
\begin{tabular}{l|ccccc|cccc}
   \toprule
   & \multicolumn{5}{c|}{PartNet Test Set} & \multicolumn{4}{c}{ShapeNet Test Set} \\
\midrule
\multirow{2}{*}{Combination} & Suc. Rate  & Mid. Error   & Head. Error    & Rot. Error   & Contact Ratio  & Suc. Rate  & Mid. Error   & Head. Error    & Rot. Error  \\
         & [\%] $\uparrow$ & [cm] $\downarrow$ & [rad] $\downarrow$ & [rad] $\downarrow$ & [\%] $\uparrow$ & [\%] $\uparrow$ & [cm] $\downarrow$ & [rad] $\downarrow$ & [rad] $\downarrow$ \\
\midrule
Direction    & \textbf{96.1} & -     & \textbf{0.256} & -    & \textbf{90.2} & \textbf{82.5} & - & \textbf{0.268} & - \\
Direction+Rotation  & 95.1 & -     & 0.263 & \textbf{0.303} & 87.9 & 81.8 & - & 0.276 & \textbf{0.320} \\
Direction+Midpoint  & 95.2 & \textbf{2.84} & 0.268 & -    & 88.7 & 81.4 & \textbf{3.15} & 0.284 & - \\
\midrule
Dir.+Rot.+Mid. & 95.0 & 2.85 & 0.270 & 0.306 & 86.7 & 81.0 & 3.22 & 0.292 & 0.338 \\
   \bottomrule
\end{tabular}
}
\vspace{-3mm}
\end{table}

\myparagraph{Evaluation on Different Objective Combinations}
Our method's adaptability was tested across various objective combinations to gauge performance impacts. Evaluations were conducted for different scenarios: solely controlling heading direction ($\bar{\V{m}}$), adding wrist rotation to heading direction ($\bar{\V{v}} + \bar{\omega}$), combining heading direction with midpoint position ($\bar{\V{v}} + \bar{\V{m}}$), and integrating all three objectives ($\bar{\V{v}} + \bar{\omega} + \bar{\V{m}}$). This variety helps to understand our approach's efficacy in handling multiple, concurrent objectives and understanding the complexity levels our method can effectively manage.
For each objective set, we sample 25 sets of motion objectives for PartNet objects and 5 for ShapeNet objects. Results in \reftab{tab:individual} show slightly better performance with fewer motion objectives, highlighting the increased control challenge with multiple objectives. This demonstrates our method's capability to handle varying difficulty levels and its robustness, as indicated by the minimal performance decline.

\section{Conclusion}
We presented \method, an RL-based method that learns to synthesize grasping motions on large-scale while satisfying one or multiple motion objectives. We introduced a learning curriculum to deal with the control complexity and an objective-driven guidance technique to accelerate exploration during training. To improve the generalization ability, we adopted a joint distance sensor to capture object local shape features. Notably, we demonstrated that our method exhibits a high success rate of 82.2\% on a test set with more than 500k unseen objects. Moreover, we showed that our approach works well even when applied to different dexterous hand platforms and with reconstructed or generated objects.

\bibliographystyle{splncs04}
\bibliography{egbib}

\clearpage
{\Large \center
\textbf{{GraspXL: Generating Grasping Motions for Diverse Objects at Scale}}\\
\vspace{0.5em}Supplementary Material \\
\vspace{1.0em}
}

In \refsec{supp:qualitative}, we provide qualitative results compared against our baseline. 
We then provide the implementation details about hyperparameters and motion objectives in \refsec{supp:impl}. In \refsec{supp:exp_details}, we show the experiment details about data preprocessing and the evaluation with generated and reconstructed objects. Finally, we provide additional experiments in \refsec{supp:experiments}.
Our model, code, and the large-scale generated motions are released for future research. Check our project page: \href{https://eth-ait.github.io/graspxl/}{https://eth-ait.github.io/graspxl/} for more details and visualization.

\begin{figure}[b]
  \centering
  \includegraphics[width=0.99\linewidth]{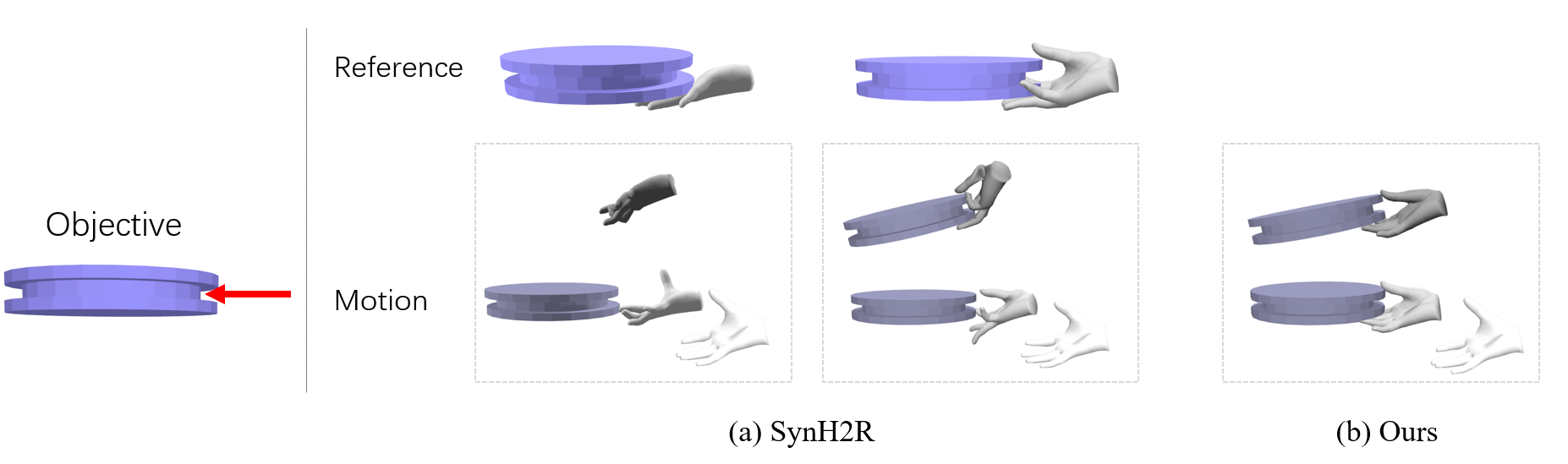}
  \caption{\textbf{Qualitative Comparison.} SynH2R requires a time-consuming reference generation process, and suffers from noisy references and imperfect reference tracking, which lead to failed grasping or large objective errors.}
\label{fig:qualitative}
\end{figure}

\section{Qualitative Results}
\label{supp:qualitative}
We provide qualitative comparisons of our method with SynH2R~\cite{christen2023synh2r} for objective-driven grasping synthesis in~\reffig{fig:qualitative}. From the figures, we can see that SynH2R either failed to establish a stable grasp due to noisy generated references, or cannot precisely follow the objectives. However, our method does not require a pre-generated reference, and can generate motions with stable grasping while satisfying motion objectives.

\section{Implementation Details}
\label{supp:impl}
\subsection{Training Hyperparameters}
We use PPO~\cite{schulman2017proximal} to train our policy and follow the implementation provided in \cite{christen2022dgrasp}. We present an overview of the important parameters and weight values of the reward function in \reftab{tab:params} and \reftab{tab:weight}. 

\begin{table}[ht]
\caption{\textbf{Hyperparameters of \method.}}

\centering
\resizebox{0.45\columnwidth}{!}{
\begin{tabular}{ll}
\toprule
\textbf{Hyperparameters PPO} & \textbf{Value} \vspace{0.1cm}\\
\midrule\\
Epochs & 1e4\\
Steps per epoch & 3e4\\
Environment steps per episode & 150 \\
Batch size & 2000 \\
Updates per epoch & 20 \\
Simulation timestep & 2.5e-3s \\
Simulation steps per action & 4 \\
Discount factor $\gamma$ & 0.996 \\
GAE parameter $\lambda$ & 0.95 \\
Clipping parameter &0.2 \\
Max. gradient norm & 0.5 \\
Value loss coefficient & 0.5\\
Entropy coefficient & 0.0\\
Optimizer & Adam\\
Learning rate & 5e-4\\
Hidden units & 128 \\
Hidden layers & 2 \\
\bottomrule
\end{tabular}
}
\label{tab:params}
\end{table}
\begin{table}[ht]
\caption{\textbf{Weights of the Reward Function.}}

\centering
\resizebox{0.5\columnwidth}{!}{
\begin{tabular}{lll}
\toprule
\textbf{Weights} & \textbf{Value (1st phase)} & \textbf{Value (2nd phase)}\\
\midrule
$w_{d}^{+}$ & 0.3 & 0.3\\
$w_{d}^{-}$ & 0.06 & 0.06\\
$w_{\V{v}}$ & 1.0 & 0.01\\
$w_{\omega}$ & 1.0 & 0.01\\
$w_{\V{m}}$ & 10.0 & 10.0\\
$w_{c}^{+}$ & 1.0 & 1.0\\
$w_{c}^{-}$ & 1.0 & 1.0\\
$w_{f}^{+}$ & 0.3 & 0.5\\
$w_{f}^{-}$ & 0.15 & 0.25\\
$w_{h}$ & 0.001 & 0.001\\
$w_{o}$ & 0.0 & 0.1\\
$w_{anatomy}$ & 0.2 & 0.1\\
$\lambda$ & 5.0 & 5.0\\
\bottomrule
\end{tabular}
}
\label{tab:weight}
\end{table}

\subsection{Objectives Specification}
As explained in Section 3 of the main manuscript, our framework can deal with different combinations of four kinds of objectives: the partition of graspable/non-graspable object point cloud $\{\V{o}^+_j\} \cup \{\V{o}^-_j\}$, the heading direction of the hand $\bar{\V{v}}$, the hand wrist rotation $\bar{\omega}$, and the hand midpoint position $\bar{\V{m}}$. 
We assume the partition $\{\V{o}^+_j\} \cup \{\V{o}^-_j\}$ is specified by a user to indicate the desired grasping area, such as a mug handle or a headphone earcup. By default, $\{\V{o}^+_j\} = \{\V{o}_j\}$ and $\{\V{o}^-_j\} = \emptyset$, which means that the hand can grasp the entire object. $\bar{\V{v}}$ is the only quantity that is mandatory to specify.
$\bar{\omega}$ is by default $0$ so that the y-axis of the hand local coordinate system (See Fig. 4 in the main manuscript) is parallel to the narrowest edge of the $\{\V{o}_j\}$ projection along $\bar{\V{v}}$, which represents the easiest setting for grasping with the given heading direction $\bar{\V{v}}$. $\bar{\V{m}}$ is by default set to be the centroid of $\{\V{o}_j\}$.

\section{Experimental Details}
\label{supp:exp_details}
\subsection{Dataset Preprocessing}

In order to compare with existing methods and to demonstrate our method's generalization capabilities, we use the  three object datasets: PartNet~\cite{partnet2019}, ShapeNet~\cite{shapenet2015}, and Objaverse~\cite{objaverse}. Since not all objects are feasible for grasping (such as a piece of paper), we preprocess and filter the objects.

\textbf{PartNet} We select 80 objects from the categories scissors, knife, mug, earphone, and wineglass, which contain part-based segmentation  
(such as the handle and the main body of a mug).
We use the individual parts of each object to specify the graspable/non-graspable areas. 
Objects are resized to feasible dimensions for grasping. 

\textbf{ShapeNet} We utilize the objects of ACRONYM~\cite{acronym2020} (scaled ShapeNet~\cite{shapenet2015} objects) and filter out extreme-sized objects. Specifically, we remove objects with a minimal bounding box width larger than 0.1$m$ or smaller than 0.01$m$, or maximal bounding box width larger than 0.3$m$, or a volume smaller than 8$cm^3$. This leads to 4019 objects.

\textbf{Objaverse} To show generalization across different scales, we resize the Objaverse~\cite{objaverse} objects to three different scales: small, medium, and large. Specifically, for each object, we uniformly sample a small scale $s\in[3,5]cm$, a medium scale $m\in[5,7]cm$, and a large scale $l\in[7,9]cm$. We then resize each object to three so that the minimum dimension of their bounding box is equal to $s$, $m$, and $l$, accordingly. Finally, we remove the objects with a maximal bounding box width larger than 0.3$m$ or smaller than 0.05$m$. This leads to 503,409 objects. 

As the object meshes from the datasets contain no material information for density and friction, we calculate object masses based on the mesh volume for a given fixed density, leading to diverse masses. We use the same friction coefficient in simulation for all objects. 

\subsection{Reconstructed and Generated Objects}
We use the eight objects generated with DreamFusion~\cite{poole2022dreamfusion} which are available from their project page, and manually scale them to graspable sizes as our test set for reconstructed objects. For reconstructed objects, we use all the fourteen objects reconstructed from HO3D~\cite{hampali2020honnotate} videos and six objects reconstructed from in-the-wild videos reported in HOLD\cite{fan2024hold}. For comparison, we evaluate with the ground-truth HO3D objects with their original scales. For each object, we randomly sample 25 sets of motion objectives and report the average performance.

\section{Additional Experiments}
\label{supp:experiments}
\subsection{Training Set Size Effect Evaluation}
To show the data efficiency of our method, we train another policy with an enlarged training set composed of 100 PartNet objects (with the graspable area partitions), and 400 ShapeNet objects. We then perform the same evaluation with the PartNet and ShapeNet test sets (See Section 4.1 in the main manuscript). The results are shown in~\reftab{tab:train}. Compared with the results with a smaller training set (See Section 4.2 in the main manuscript), there is no significant improvement, which means that a training set composed of 58 objects with diverse shape is sufficient for our framework. This shows the data efficiency of our method. We hypothesize that this is because diverse shapes together with random objectives and initialization during training provide a diverse distribution of grasps and configurations.

\begin{table}
\centering
\caption{
\textbf{Comparison with Different Training Set Size.}
}
\label{tab:train}
\resizebox{\columnwidth}{!}{
\begin{tabular}{l|ccccc|cccc}
   \toprule
   & \multicolumn{5}{c|}{PartNet Test Set} & \multicolumn{4}{c}{ShapeNet Test Set} \\
\midrule
\multirow{2}{*}{Method} & Suc. Rate  & Mid. Error   & Head. Error    & Rot. Error   & Contact Ratio  & Suc. Rate  & Mid. Error   & Head. Error    & Rot. Error  \\
         & [\%] $\uparrow$ & [cm] $\downarrow$ & [rad] $\downarrow$ & [rad] $\downarrow$ & [\%] $\uparrow$ & [\%] $\uparrow$ & [cm] $\downarrow$ & [rad] $\downarrow$ & [rad] $\downarrow$ \\
\midrule
Ours (58 Training Objects)  & \textbf{95.0} & \textbf{2.85} & \textbf{0.270} & 0.\textbf{306} & 86.7 & 81.0 & \textbf{3.22} & 0.292 & \textbf{0.338} \\
Ours+ (500 Training Objects) & 94.9 & 3.57 & 0.307 & 0.356 & \textbf{88.1} & \textbf{84.3} & 3.94 & \textbf{0.283} & 0.346 \\
   \bottomrule
\end{tabular}
}
\end{table}

\subsection{Friction Effect Evaluation}
Although we use the same friction coefficient in simulation for all objects, our method can also deal with randomized frictions (±0.3 around the default friction coefficient) as shown in~\reftab{tab:friction}.

\begin{table}
\centering
\caption{
\textbf{Evaluation with random friction coefficients.}
}
\label{tab:friction}
\resizebox{0.7\columnwidth}{!}{
\begin{tabular}{l|cc}
   \toprule
   Settings & Suc. Rate [\%] $\uparrow$ & Mid. Error $\downarrow$ \\
\midrule
Original experiment & 81.6 & 0.283 \\
Random friction coefficient & 80.5 & 0.285 \\
   \bottomrule
\end{tabular}
}
\end{table}

\subsection{Realism Evaluation}
To evaluate the realism of our method, we invite 35 participants to score the realism in terms of human likeness, naturalness, smoothness, and hand-object interpenetration. They score 10 sets of rendered grasping motions from 1 (worst) to 3 (best), each containing three motions of the same object randomly chosen from ours, HO3D~\cite{hampali2020ho3d}, and DexYCB~\cite{chao2021dexycb}. The average scores are 2.12, 2.05, and 2.45, respectively. Our method has a slightly higher realism score than HO3D as HO3D has some interpenetrations and jitters caused by labeling noise, and DexYCB has the highest score due to accurate annotations.

\subsection{Inference Speed Evaluation}
With a 24-core Intel i9-14900 CPU and an Nvidia RTX 4090 GPU, we generate 100 sequences and calculate the average time consumption per sequence with our method, the baselines SynH2R-PD, and SynH2R~\cite{christen2023synh2r}, leading to 37.15, 37.17, and 0.14 seconds, respectively. The most time consumption for SynH2R-PD and SynH2R are caused by the optimization-based static grasping reference pose generation procedure.

\end{document}